\documentclass{article}

\usepackage{microtype}
\usepackage{graphicx}
\usepackage{subcaption}
\usepackage{booktabs}

\usepackage{hyperref}
\usepackage{xurl}

\usepackage[preprint]{icml2026}

\usepackage{amsmath}
\usepackage{amssymb}
\usepackage{mathtools}

\usepackage[capitalize,noabbrev]{cleveref}
\usepackage{placeins}

\icmltitlerunning{Compressed Computation under \smash{$L^4$} Loss is likely Computation in Superposition}

\begin{document}

\twocolumn[
  \icmltitle{Compressed Computation under $L^4$ Loss is likely Computation in Superposition}

  \icmlsetsymbol{equal}{*}

  \begin{icmlauthorlist}
    \icmlauthor{Francisco Ferreira da Silva}{pivotal}
    \icmlauthor{Stefan Heimersheim}{}
  \end{icmlauthorlist}

  \icmlaffiliation{pivotal}{Pivotal Research}

  \icmlcorrespondingauthor{Francisco Ferreira da Silva}{franhfsilva@gmail.com}

  \icmlkeywords{Machine Learning, ICML, superposition, computation in superposition, interpretability, toy models}

  \vskip 0.3in
]

\printAffiliationsAndNotice{}

\begin{abstract}
Neural networks are thought to represent concepts as directions in their activation space, and superposition lets them encode more concepts than they have dimensions.
It is natural to ask whether they can also compute more functions than they have neurons, i.e., perform computation in superposition.
In this regime many functions of sparse inputs are evaluated by a layer with fewer neurons than there are functions to compute.
Representation in superposition is by now fairly well understood, but computation in superposition is not, and there are few toy models of it arising through training rather than being hand designed.
As a toy model of computation in superposition we study the compressed-computation setup: a single-hidden-layer ReLU network with 50 neurons that must compute the ReLU of each of 100 sparse input features.
We show that training it under an $L^4$ loss (the mean fourth power of the error), rather than the usual $L^2$, elicits a solution that appears to compute all features in superposition.
We then reverse-engineer this solution.
We find that the network assigns each feature a sparse binary codeword over neurons and decodes it with a pseudoinverse of the encoder.
Given these codewords, a description with only three scalars recovers most of the network's performance, and we validate it by building equivalent networks from hand-designed codes.{\renewcommand{\thefootnote}{$\dagger$}\footnote{Source code available at \url{https://github.com/FranciscoHS/toy-model-cis-code}.}}
\end{abstract}

\section{Introduction}
Neural networks are widely assumed to make use of \textit{superposition} to represent more concepts than they have dimensions~\cite{elhage2022toy}.
A natural follow-up is whether they also \textit{compute} in superposition, i.e., implement more non-linear functions than they have dimensions~\cite{hanni2024mathematical,adler2024complexity}.

While representation in superposition is increasingly well understood, computation in superposition (CiS) is not.
In particular, until very recently~\cite{gibson2026bloom} there were no toy models of computation in superposition \textit{in the wild}, i.e., arising through training rather than being hand designed.
Such toy models are desirable: they allow studying CiS at a scale where we can hope to fully reverse engineer the trained networks.

\citet{braun2025apd} introduced the toy model of compressed computation, where a single-hidden-layer network must compute more ReLUs than its hidden width.
However, \citet{bhagat2025compressed} showed that this toy model did not actually elicit computation in superposition.

In this work, we train a version of the compressed computation toy model under $L^4$ loss rather than $L^2$ and argue that the resulting network performs computation in superposition.

Our contributions are:

\begin{enumerate}
  \item We train a toy model of compressed computation with $L^4$ loss and argue that it performs computation in superposition.
  \item We reverse-engineer the trained network: it assigns each feature a sparse binary codeword over neurons and decodes with a pseudoinverse of the encoder. Given these codewords, a 3-scalar parameterization (on-code value, off-code value, decoder scale) recovers $\approx 1.1\times$ its loss.
  \item We validate this description by substituting hand-designed binary codes in the same 3-parameter family, yielding equivalent networks at $\approx 1.1\text{-}1.2$x the trained model's loss.
\end{enumerate}
\section{Related Work}
\paragraph{Comparable empirical work.}
\citet{gibson2026bloom} is concurrent work similar in spirit to ours: both partially reverse-engineer one-layer ReLU networks that learn binary codes, and the aspects that remain unexplained are nearly identical --- non-uniform on-codeword values, a decoder that deviates from a scaled pseudoinverse, and a trained network that outperforms designed alternatives.
The key difference is the task. Both networks take continuous sparse inputs, but ours has a regression task (compute the ReLU of each input), whereas Gibson's network must identify which inputs are active, outputting a fixed value at each active index regardless of its magnitude.

\paragraph{Theoretical constructions for CiS.}
~\citet{gibson2025pingpong} considers a task in which a continuous input passes through one of many small circuits sharing a single wider network, with exactly one circuit active per forward pass.
They provide a hand-designed zero-error construction for computation in superposition via sparse memory blocks; we find a sparse-coding solution arising through training.

\paragraph{Random sparse binary codes.}
\citet{hanni2024mathematical} proposes random sparse binary codes for computation in superposition.
Our work appears to be an empirical instance of their constructions.

\paragraph{The compressed computation toy model.}
\citet{braun2025apd} introduced the toy model of compressed computation, and~\citet{bhagat2025compressed} showed that it does not elicit computation in superposition.
We train a version of this model under $L^4$ loss and show that it does.

\section{Methodology}
\label{sec:methodology}
\begin{figure}[ht]
  \centering
  \includegraphics[width=\linewidth]{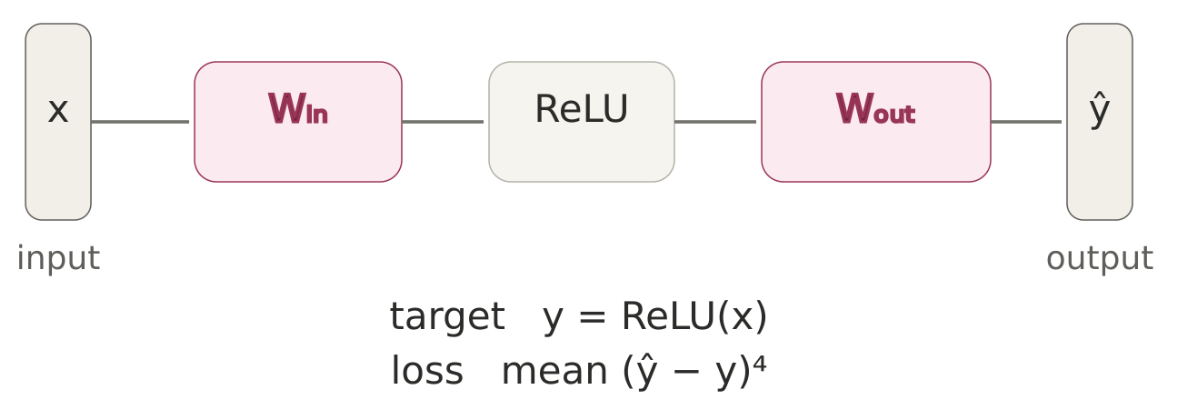}
  \caption{\textbf{Network architecture:} a linear encoder $W_{\rm in}$, 50 ReLU neurons without bias, and a linear decoder $W_{\rm out}$. The input and output are 100-dimensional; the task is to compute the elementwise ReLU of the input.
  }
  \label{fig:architecture}
\end{figure}

Our network (\cref{fig:architecture}) is the compressed-computation toy model of \citet{braun2025apd} with the residual connection removed (see below).
The input is $100$-dimensional, with each feature being zero with probability $1-p$ and drawn uniformly from $(-1,1)$ with probability $p$; we set $p=0.02$, giving two active features in expectation.
A single hidden layer of $50$ ReLU neurons without bias acts directly on the input through trainable weights $W_{\rm in} \in \mathbb{R}^{50 \times 100}$ and $W_{\rm out} \in \mathbb{R}^{100 \times 50}$, and the task is to compute the elementwise ReLU of the input.
\citet{braun2025apd} additionally embed the input as random near-orthogonal directions in a higher-dimensional ($1000$) space through a fixed random matrix $W_E$ with unit-norm rows, read back out by the transpose $W_E^\top$, so that the features are not axis-aligned --- the more realistic and challenging setting for the parameter-decomposition methods the model was built to test.
That embedding turns out to be incidental to the computation-in-superposition solution: training with it recovers an identical solution.
We therefore present the simpler axis-aligned model in the main text and defer the embedded variant to \cref{app:embedding}.
Beyond setting aside the incidental embedding, we also remove the residual connection, motivated by \citet{bhagat2025compressed}'s analysis.
They show that, in Braun et al.'s model, the residual connection is equivalent to a mixing matrix $M$ in the target, $y = \text{ReLU}(x) + Mx$, where $M$ couples input $x_j$ to target $y_i$ for $i \neq j$ --- and that this $M$, not superposition, is what lets the model beat the naive baseline of ignoring half the features.
Removing the residual connection sets $M = 0$, so the target is exactly $y = \text{ReLU}(x)$ and any substantial improvement over the naive baseline must come from superposition.
This removal is necessary but not sufficient: \citet{bhagat2025compressed} showed that dropping the residual alone does not elicit computation in superposition, and we confirm it --- trained under the usual $L^2$ loss our model still finds the naive solution (\cref{sec:evidence}).

The change we make that elicits CiS is the loss.
We train under $L^4$ loss, the average fourth power of the elementwise error, rather than the usual $L^2$.
We train with Adam (learning rate $0.01$, cosine annealing schedule), batch size $8{,}192$, for $100{,}000$ steps; $W_{\rm in}$ and $W_{\rm out}$ are initialized entrywise from $U(-0.1, 0.1)$ and $U(-0.15, 0.15)$.
In \cref{sec:evidence} we also sweep the loss exponent, including standard $L^2$.
The intuition behind making this change is that $L^4$ penalizes outlier errors more than $L^2$, and hence should prioritize networks approximating all features roughly equally well (superposition) rather than computing the output for some features perfectly and for others not at all.

\section{Results}
\subsection{Evidence for computation in superposition}
\label{sec:evidence}
\begin{figure}[ht]
  \centering
  \includegraphics[width=\linewidth]{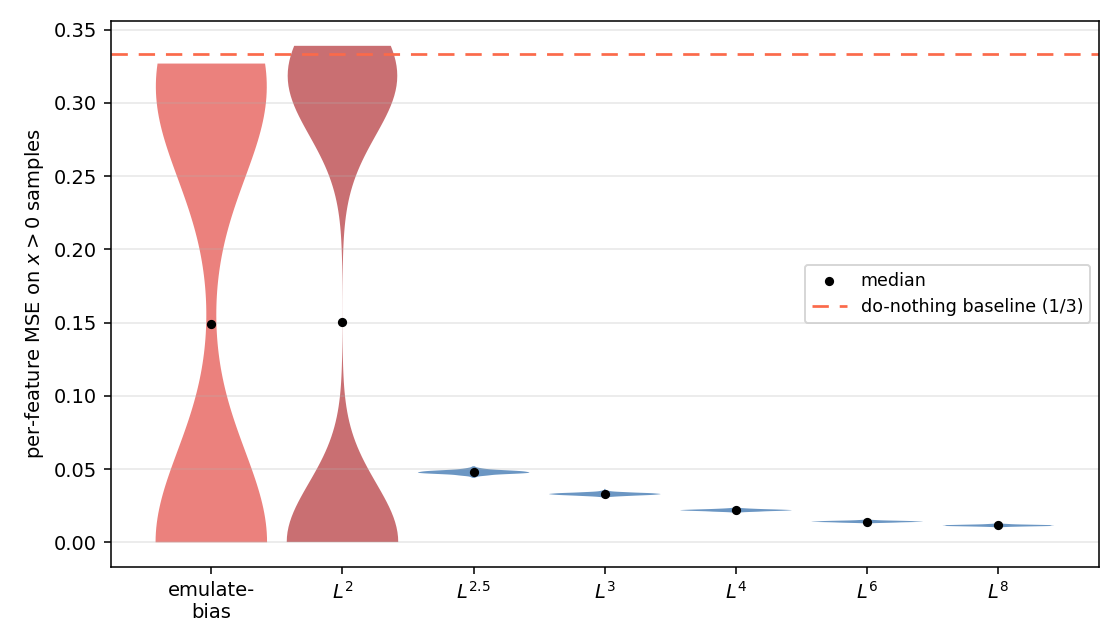}
  \caption{\textbf{Per-feature MSE on $x>0$ samples: the strongest non-superposition baseline and the trained networks at several loss exponents.} Each violin pools the per-feature MSE (over the $100$ features, and over $10$ seeds for the trained networks); black dots are medians. The emulate-bias baseline and $L^2$ (reds) are bimodal --- half the features handled, half left near the do-nothing baseline of $1/3$ (dashed), which the bias offset lowers only marginally. Every exponent above $2$ (blue) instead concentrates the error into a narrow band far below the baselines, spread evenly across all $100$ features.}
  \label{fig:per-feature-loss}
\end{figure}

As evidence that $L^4$ training elicits superposition, we compare the per-feature loss of the $L^2$- and $L^4$-trained networks (\cref{fig:per-feature-loss}).
Training under $L^2$ results in a ``naive solution'' that learns half the features perfectly and ignores the rest (red).
Under $L^4$, the trained network instead learns all 100 features approximately equally well.
We further compare against another no-superposition baseline, namely one that represents half the features exactly and emulates a bias on the rest by adding a constant offset to their outputs (following \citet{bhagat2025compressed}).
At our sparsity the offset helps only marginally, with the emulate-bias baseline having loss $\sim\!7\%$ lower than the naive baseline, still $\sim\!26\times$ higher than the $L^4$ solution.

As there are more features than neurons, achieving low error on all features requires sharing neurons across features, i.e., superposition.
Eliciting this solution is not unique to $L^4$, as \cref{fig:per-feature-loss} shows: every exponent above $2$ produces the same concentrated distribution.
To quantify this, we summarize each network by the coefficient of variation of its per-feature MSE across the 100 features --- the standard deviation divided by the mean.
We use the coefficient of variation rather than the raw standard deviation because the mean per-feature loss varies greatly across exponents, so a scale-free measure is needed to compare them; it is small when the error is spread evenly across features (indication of superposition) and large when a few features carry most of the error (the naive solution).
Every exponent above $2$ that we tried ($2.5, 3, 4, 6, 8$) gives a superposition-like solution, with coefficients of variation between $0.028$ and $0.046$; $L^2$ instead gives the naive solution, with a coefficient of variation of $0.999$.
Across seeds the coefficient of variation varies by at most $0.003$ throughout.
$L^4$ is thus a representative choice rather than a special one; \cref{fig:per-feature-loss} also shows $L^3$ and $L^6$, which cluster with $L^4$ well below the baselines.

\subsection{The mechanism}

We characterize the $L^4$-trained network's mechanism through three observations about its encoder $W_{\rm in} \in \mathbb{R}^{N \times F}$, which maps each input feature to the neuron pre-activations it induces, and its decoder $W_{\rm out} \in \mathbb{R}^{F \times N}$.

\begin{figure}[ht]
  \centering
  \includegraphics[width=\linewidth]{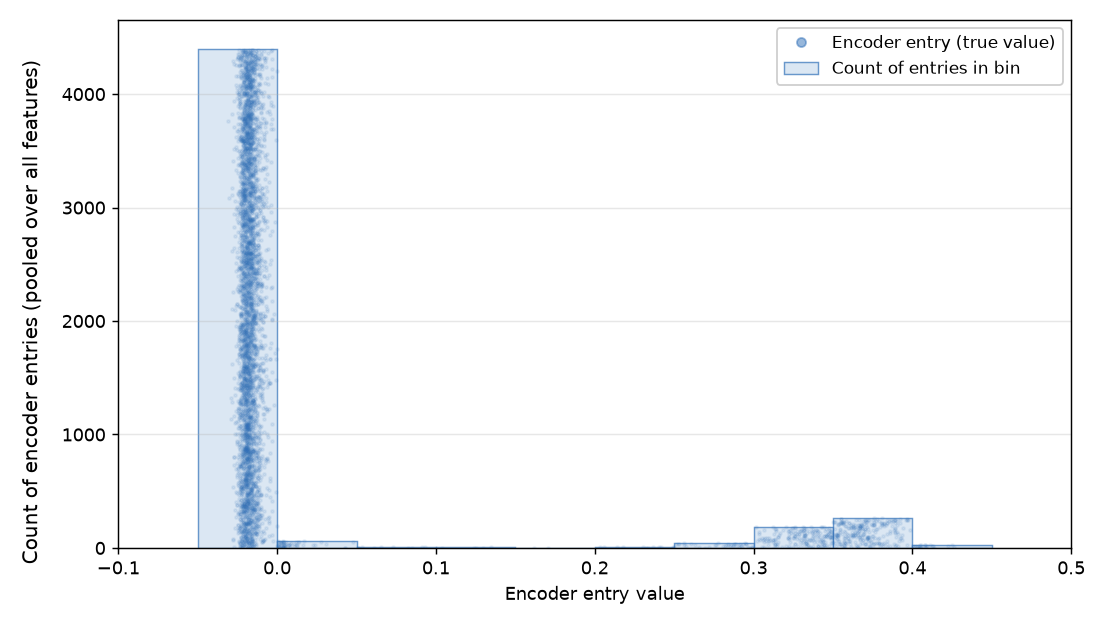}
  \caption{\textbf{Encoder entries are bimodal:} pooled histogram of all entries of the encoder $W_{\rm in}$ over the $100$ feature columns. Bars are binned in steps of $0.05$ with edges aligned to zero; bar height is the number of entries in each bin, with those entries scattered over the bar at their true value (vertical position is jitter, for visibility). The entries split into a large off-code cluster of small negatives ($-0.02 \pm 0.01$) and a smaller on-code cluster of large positives ($0.34 \pm 0.06$).}
  \label{fig:value-hist}
\end{figure}

\Cref{fig:value-hist} shows the distribution of the entries of the encoder $W_{\rm in}$, pooled over all $100$ feature columns.
The entries fall into two well-separated groups (values reported as mean $\pm$ standard deviation): a large mass of small negative values ($-0.02 \pm 0.01$) and a smaller mass of large positive values ($0.34 \pm 0.06$).
We call the support of the large entries the \textit{codeword} of feature $j$, written $M_j \in \{0,1\}^N$ --- the subset of neurons that fire when feature $j$ is active, with codeword length $K = |M_j|$.
Stacking these gives a binary matrix $M \in \{0,1\}^{F\times N}$ whose pattern matches the shape of $W_{\rm in}$: large positives where $M_{j,n}=1$, small negatives where $M_{j,n}=0$.

\begin{figure}[ht]
  \centering
  \includegraphics[width=\linewidth]{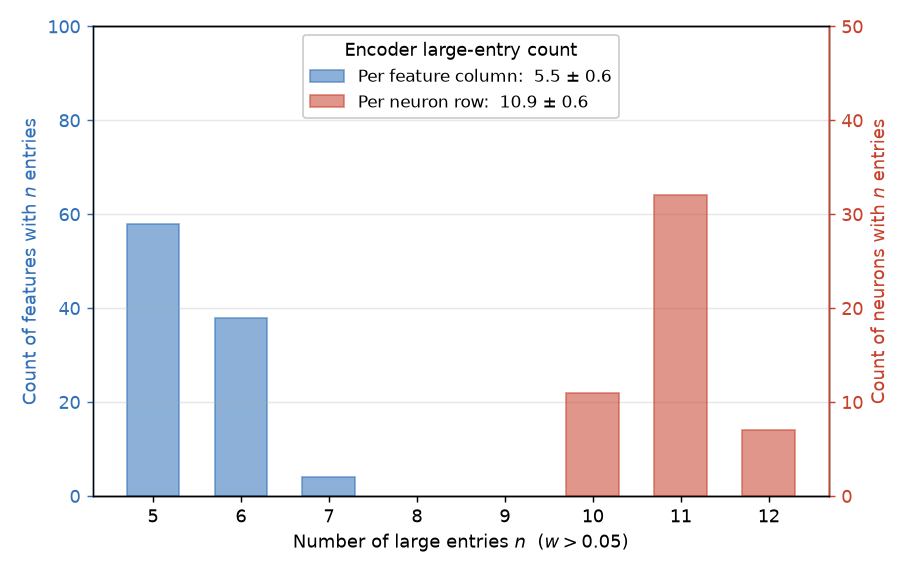}
  \caption{\textbf{The binary code is regular:} number of large encoder entries (those exceeding $0.05$, i.e.\ on-code entries) per feature column (blue, left axis) and per neuron row (red, right axis). Every feature uses $5\text{-}7$ neurons (mean $5.5$) and every neuron is shared by $10\text{-}12$ codewords (mean $10.9$); the tight bimodal distribution shows the code is close to biregular.}
  \label{fig:code-hist}
\end{figure}

The codewords are quite regular (\cref{fig:code-hist}): every feature uses $K=5\text{-}7$ neurons (mean $5.5$), and every neuron participates in $10\text{-}12$ codewords (mean $10.9$).
The code is thus close to \emph{biregular} --- every codeword nearly the same length, and every neuron shared by nearly the same number of codewords --- so the decoding load is spread near-equally across neurons.

\begin{figure}[ht]
  \centering
  \includegraphics[width=\linewidth]{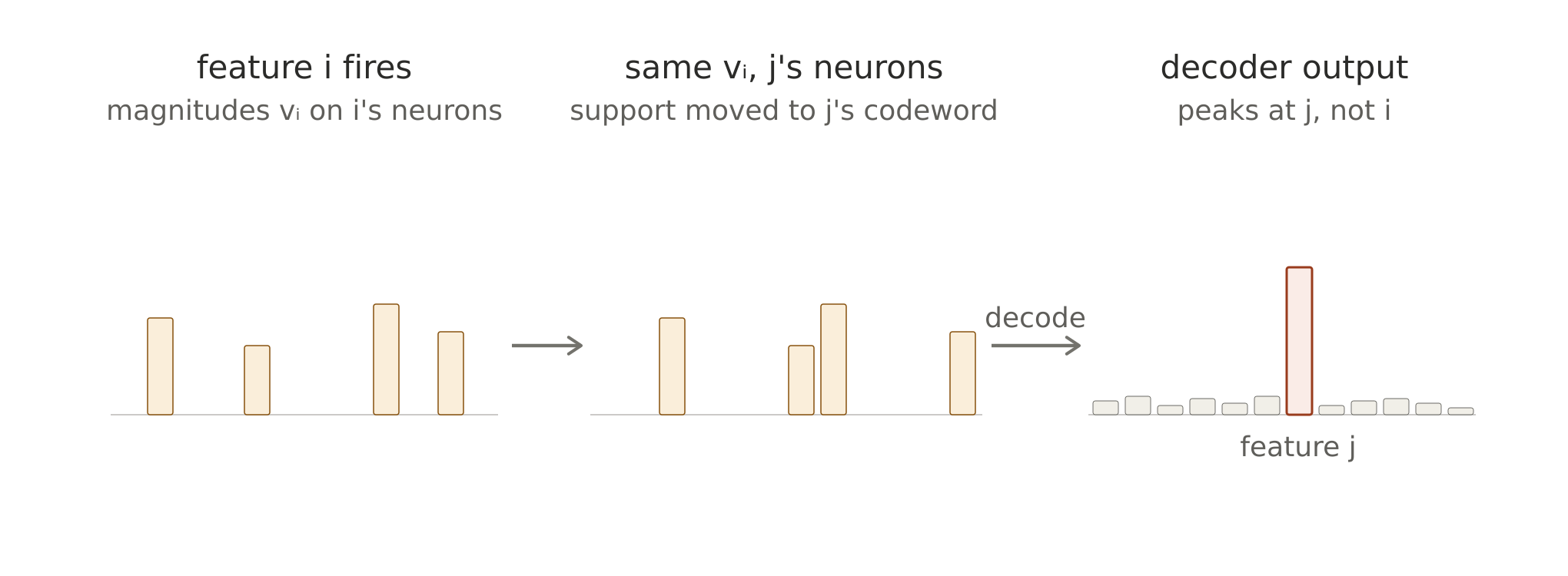}
  \caption{\textbf{Codeword swap test:} the hidden magnitudes produced by feature $i$, placed on feature $j$'s codeword neurons, decode to $j$. The codeword positions, not the magnitudes, determine the decoded feature.}
  \label{fig:swap}
\end{figure}

If the codewords serve as the network's internal identifier for each feature, then transplanting feature $i$'s hidden activation values onto feature $j$'s codeword neurons should make the decoder output feature $j$.
We verify this directly: for every pair of features $(i, j)$ of equal codeword length we transplant feature $i$'s hidden magnitudes onto feature $j$'s codeword neurons, and the decoder's top output is $j$ in $100\%$ of cases (\cref{fig:swap}).

Finally, the decoder seems to do little more than invert the encoder: the decoder $W_{\rm out}$ has Frobenius cosine $0.996$ with a scaled pseudoinverse of the encoder (the Frobenius cosine is the cosine similarity between the two flattened matrices).
To confirm, we replace the trained decoder outright with a scaled pseudoinverse of the encoder and find the $L^4$ loss increases only slightly ($1.15\times$ the trained network), so the decoder is well described as a pseudoinverse read-out of the codewords.
Because the codewords have small pairwise overlap, this pseudoinverse is itself close to a scaled transpose of the encoder, so the decoder ends up looking much like the encoder: each feature is read out along essentially the same neurons that encode it.

\begin{figure}[ht]
  \centering
  \includegraphics[width=\linewidth]{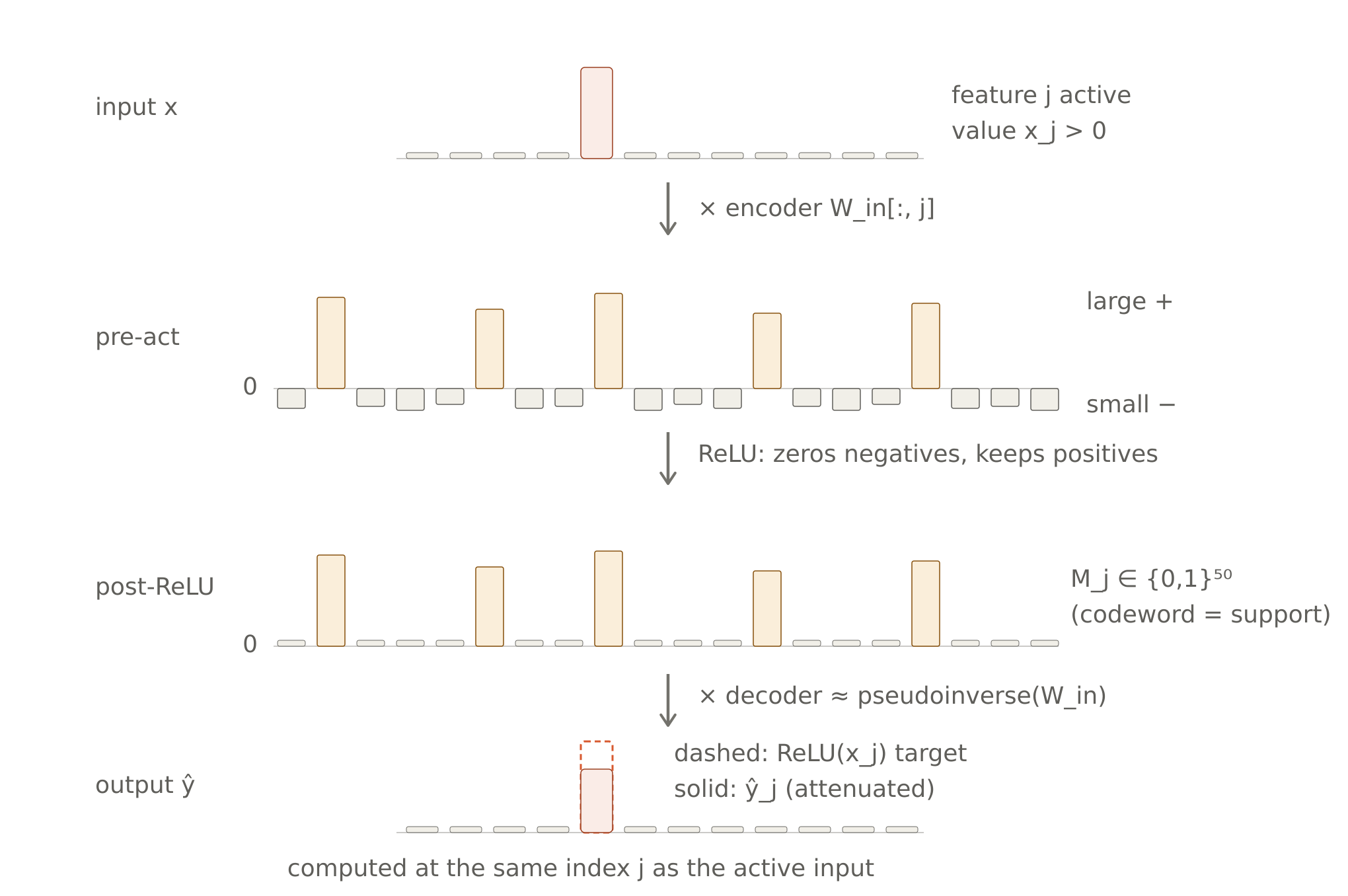}
  \caption{\textbf{Mechanism for a single active feature:} the encoder produces large positives on the feature's codeword neurons and small negatives elsewhere; ReLU zeros the negatives; the pseudoinverse decoder reads the codeword and outputs a peak at the same index.}
  \label{fig:mechanism}
\end{figure}

Together, these findings paint the picture in \cref{fig:mechanism}: a binary-code encoder followed by a pseudoinverse decoder.
One feature of this picture is that the decoded output is \emph{attenuated}: the network systematically under-shoots the true ReLU. This is a response to interference --- because codewords overlap, an active feature leaks through its shared neurons into the outputs of the (far more numerous) inactive features, and shrinking every active output keeps that leakage small. The single-feature input--output response in \cref{fig:io-response} shows this directly.

\subsection{A 3-parameter description given a binary code}
The mechanism depicted in \cref{fig:mechanism} suggests a simple ansatz.
We extract the binary codeword matrix $M$ from the trained network by classifying each encoder entry as on- or off-code, then parameterize a new encoder with one value on the support and one off it, $W_{\rm in} = a\,M^\top + b\,(1-M^\top)$, and set the decoder to a scaled pseudoinverse of it, $W_{\rm out} = c\,(W_{\rm in})^{+}$.
This leaves three free scalars: the on-code value $a$, the off-code value $b$, and the decoder scale $c$.

Fitting these three scalars to the trained network's codewords gives an $L^4$ loss $\sim\!1.13\times$ that of the trained network, far below the non-superposition baselines (naive $\sim\!28\times$, emulate-bias $\sim\!26\times$, random network $\sim\!56\times$).
\paragraph{The support carries little information.}
We test how much the specific support matters by replacing $M$ with a randomly generated codeword matrix of a given codeword length $K$. We consider two families: \textit{biregular} codes (every codeword the same length, every neuron shared by the same number of codewords) and random codes.
We try each family both as generated and after applying random edge swaps that reduce pairwise codeword overlaps (\cref{app:overlap}), expecting biregularity to balance load across neurons and uniform overlaps to minimize the worst-case interference that $L^4$ penalizes most.

\begin{figure}[ht]
  \centering
  \includegraphics[width=\linewidth]{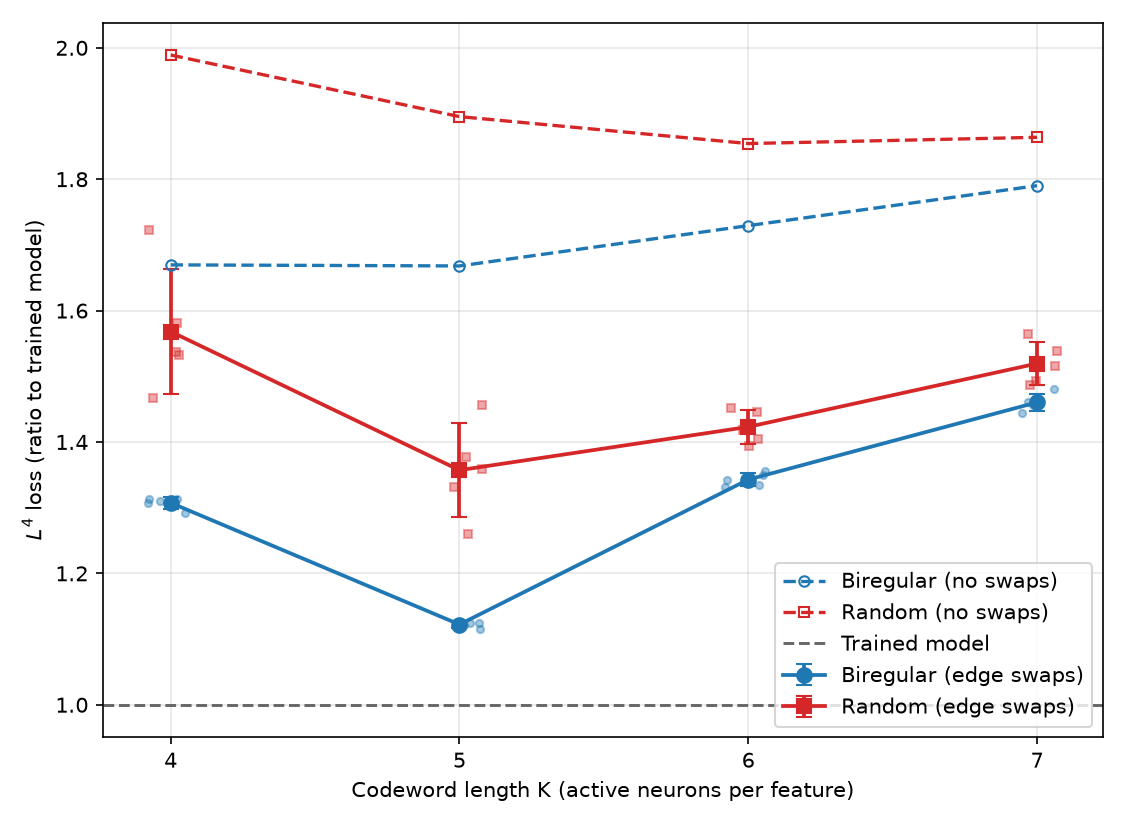}
  \caption{\textbf{$L^4$ loss of 3-parameter ansatz networks:} Loss of networks built from synthetic binary codes (biregular and random) as a function of codeword length $K$, normalized by the trained model's loss. Solid lines apply overlap-minimizing edge swaps (five seeds each; error bars are standard deviations, scatter points are individual seeds); dashed lines are the same codes without edge swaps. Biregular $K=5$ with edge swaps is best at ${\sim}1.12\times$ the trained loss; without edge swaps both families stay within ${\sim}2\times$. No synthetic code beats the trained network.}
  \label{fig:synthetic-codes}
\end{figure}

\Cref{fig:synthetic-codes} shows the result.
Overlap minimization matters: without it both families sit at $1.7\text{-}2.0\times$ the trained loss, while with it the best designed code --- biregular $K=5$ --- reaches $\sim\!1.12\times$, marginally beating the three-scalar version of the trained network's own code.
Up to the choice of code family and codeword length $K$, then, the network's performance is captured by just three scalars, and the precise support $M$ carries little extra information.

\section{Open Questions}
\label{sec:open}
The three-scalar ansatz built from the trained codewords reaches $1.13\times$ the trained loss, and the best hand-designed code (\cref{fig:synthetic-codes}) comes no closer.
Since the ansatz already uses the trained network's own support --- so the codewords, including their lengths, match --- what separates it from the trained network is just the encoder's specific per-entry values and the decoder.
\Cref{tab:impact} decomposes how much each accounts for: relaxing the encoder or the decoder closes part of the gap, and the network reaches $1.00\times$ only when both are free.
We understand each piece only partially.

\begin{table}[ht]
  \centering
  \small
  \setlength{\tabcolsep}{4pt}
  \caption{\textbf{What each ingredient accounts for.} For each construction we list the encoder, the decoder, and the resulting $L^4$ loss relative to the trained network ($1.00\times$). \emph{Trained support}: the binary codeword pattern $M$ (which neurons each feature fires), read off the trained encoder. \emph{3 scalars}: the on-code/off-code/decoder-scale ansatz $(a,b,c)$. \emph{Free per-entry}: every encoder entry fit freely. \emph{pinv}: decoder fixed to a scaled pseudoinverse of the encoder. Designed codes use codeword length $K=5$; ``overlap-min.'' applies the edge swaps of \cref{app:overlap} (\cref{fig:synthetic-codes}). The network reaches $1.00\times$ only when both encoder and decoder are free.}
  \label{tab:impact}
  \begin{tabular}{llc}
    \toprule
    Encoder & Decoder & $L^4$ loss \\
    \midrule
    \multicolumn{3}{l}{\emph{Designed code ($K=5$), 3 scalars}} \\
    \quad random, no overlap-min.    & pinv    & $1.90\times$ \\
    \quad biregular, no overlap-min. & pinv    & $1.67\times$ \\
    \quad random, overlap-min.       & pinv    & $1.36\times$ \\
    \quad biregular, overlap-min.    & pinv    & $1.12\times$ \\
    \midrule
    Trained support, 3 scalars       & pinv    & $1.13\times$ \\
    Trained support, free per-entry  & pinv    & $1.05\times$ \\
    Trained support, 3 scalars       & trained & $1.06\times$ \\
    \midrule
    Full trained encoder             & pinv    & $1.15\times$ \\
    Trained (decoder tied to pinv)   & pinv    & $1.05\times$ \\
    Full trained encoder             & trained & $1.00\times$ \\
    \bottomrule
  \end{tabular}
\end{table}

First, the per-entry encoder values.
Replacing the two on/off scalars with the trained network's individual entries, while still decoding with a scaled pseudoinverse, lowers the loss to $1.05\times$; but we do not fully understand what these values encode.
On-codeword values are not uniform, and enforcing uniformity harms performance.
We observe that neurons used in more codewords have lower on-codeword values ($r \sim -0.82$), as one would expect if the network down-weights overloaded neurons; this accounts for ${\sim}67\%$ of the variance in codeword values, so it is most but not all of the story.
Notably, the values still vary even in a designed biregular code, where every neuron has the same degree: with this load-balancing effect removed entirely, the fitted per-entry values vary about as much as in the trained network (coefficient of variation $\sim\!0.15$ versus $\sim\!0.17$).

Second, the decoder.
The trained decoder is close to a scaled pseudoinverse of the encoder (Frobenius cosine $0.996$).
A network trained \emph{from scratch} with its decoder tied to a scaled pseudoinverse of the encoder still reaches $1.05\times$, and a freely-trained decoder on the three-scalar ansatz reaches only $1.06\times$ (\cref{tab:impact}).
The larger $1.15\times$ cost of swapping in a pseudoinverse \emph{post hoc} is therefore mostly because the trained encoder was tuned for its own decoder, not because a pseudoinverse decoder is inherently worse: when the encoder co-adapts, the constraint is nearly free.

Finally, it is unclear whether the trained network's binary support is special at all: in the three-scalar family it is not, as a designed biregular $K=5$ code slightly outperforms the three-scalar version of the trained code.

Together, these results account for most of the trained network's performance and pin down its main mechanism --- a sparse binary code decoded by a near-pseudoinverse --- leaving the encoder's specific per-entry values and the decoder's small residual structure as the pieces we have not fully explained.
\section{Discussion and Future Work}

Our reverse-engineered network looks like an empirical instance of the random sparse binary codes that \citet{hanni2024mathematical} propose for computation in superposition: through training it independently arrives at the sparse-code-over-neurons motif their constructions predict.
The match is suggestive rather than exact, as their analysis targets Boolean U-AND while our network computes continuous ReLUs.

The original motivation for this toy model was a testbed for parameter-decomposition methods such as APD~\cite{braun2025apd} and SPD~\cite{bushnaq2025spd}.
Now that we have elicited superposition and reverse-engineered the network, we can and will use it as a ground-truth solution against which to test these and other interpretability methods.

More broadly, we would like to know how far this picture extends: to deeper and more realistic networks, and to more realistic kinds of computation than the elementwise ReLU studied here.
Does superposition still arise once there are several layers and bottlenecks, can we still reverse-engineer the resulting solution, and does anything resembling the error-correction layers of \citet{hanni2024mathematical} appear?

\bibliography{paper}

@article{braun2025apd,
  title={Interpretability in Parameter Space: Minimizing Mechanistic Description Length with Attribution-based Parameter Decomposition},
  author={Braun, Dan and Bushnaq, Lucius and Heimersheim, Stefan and Mendel, Jake and Sharkey, Lee},
  journal={arXiv preprint arXiv:2501.14926},
  year={2025},
  url={https://arxiv.org/abs/2501.14926}
}

@article{bushnaq2025spd,
  title={Stochastic Parameter Decomposition},
  author={Bushnaq, Lucius and Braun, Dan and Sharkey, Lee},
  journal={arXiv preprint arXiv:2506.20790},
  year={2025},
  url={https://arxiv.org/abs/2506.20790}
}

@article{hanni2024mathematical,
  title={Mathematical Models of Computation in Superposition},
  author={H{\"a}nni, Kaarel and Mendel, Jake and Vaintrob, Dmitry and Chan, Lawrence},
  journal={arXiv preprint arXiv:2408.05451},
  year={2024},
  url={https://arxiv.org/abs/2408.05451}
}

@misc{gibson2026bloom,
  title={Neural Networks learn Bloom Filters},
  author={Gibson, Alex},
  year={2026},
  howpublished={\url{https://www.lesswrong.com/posts/buxBdp8NtHGgBwabv/neural-networks-learn-bloom-filters}},
  note={LessWrong}
}

@misc{gibson2025pingpong,
  title={Ping pong computation in superposition},
  author={Gibson, Alex},
  year={2025},
  howpublished={\url{https://www.lesswrong.com/posts/g9uMJkcWj8jQDjybb/ping-pong-computation-in-superposition}},
  note={LessWrong}
}

@inproceedings{bhagat2025compressed,
  title={Compressed Computation is (probably) not Computation in Superposition},
  author={Bhagat, Jai and Molas-Medina, Sara and Giglemiani, Giorgi and Heimersheim, Stefan},
  booktitle={Mechanistic Interpretability Workshop at NeurIPS 2025},
  year={2025}
}

@article{elhage2022toy,
  title={Toy models of superposition},
  author={Elhage, Nelson and Hume, Tristan and Olsson, Catherine and Schiefer, Nicholas and Henighan, Tom and Kravec, Shauna and Hatfield-Dodds, Zac and Lasenby, Robert and Drain, Dawn and Chen, Carol and others},
  journal={arXiv preprint arXiv:2209.10652},
  year={2022}
}

@article{adler2024complexity,
  title={On the complexity of neural computation in superposition},
  author={Adler, Micah and Shavit, Nir},
  journal={arXiv preprint arXiv:2409.15318},
  year={2024}
}
\bibliographystyle{icml2026}

\appendix
\section{The embedded model}
\label{app:embedding}

\begin{figure}[ht]
  \centering
  \includegraphics[width=\linewidth]{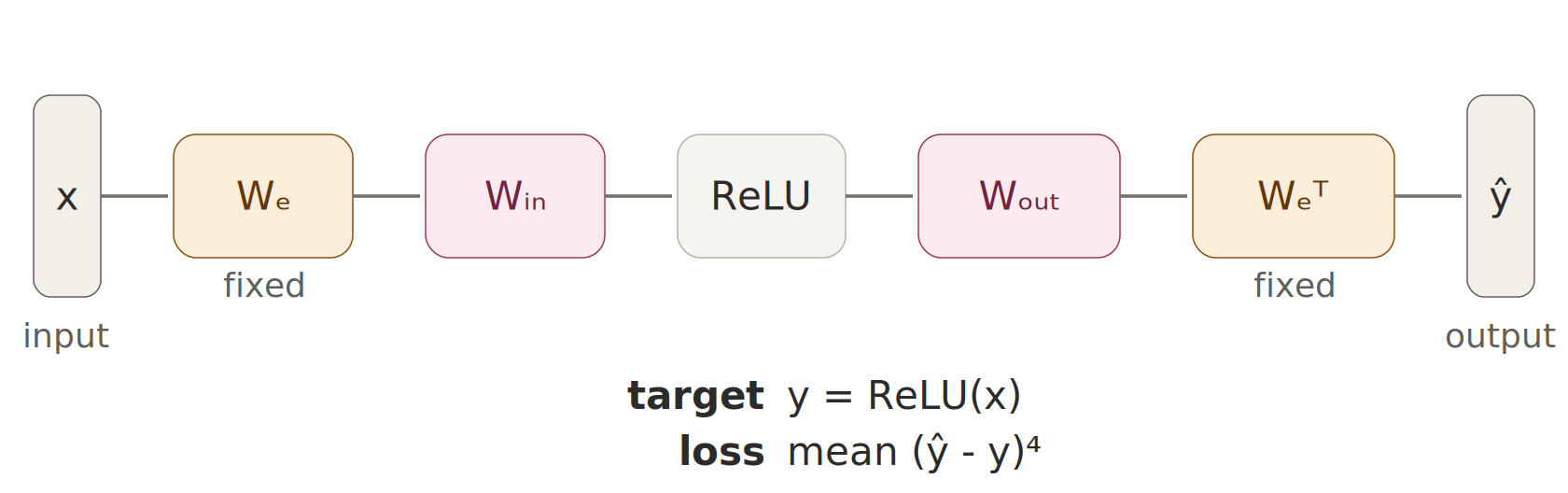}
  \caption{\textbf{Embedded architecture} (cf.\ \cref{fig:architecture}). The input is embedded as random near-orthogonal directions in a $1000$-dimensional space through a fixed matrix $W_E$ with unit-norm rows and read back out by the transpose $W_E^\top$; the trainable encoder $W_{\rm in}$, $50$ ReLU neurons, and decoder $W_{\rm out}$ are unchanged. The embedding ($W_E$, $W_E^\top$) is held fixed during training.}
  \label{fig:embed-arch}
\end{figure}

The main text studies a model in which the $50$ ReLU neurons act directly on the $100$ input features.
\citet{braun2025apd}'s original compressed-computation model instead embeds the $100$-dimensional input into $\mathbb{R}^{1000}$ through a fixed random matrix $W_E$ with unit-norm rows: the $50$ neurons read and write this $1000$-dimensional space through trainable $W_{\rm in} \in \mathbb{R}^{50 \times 1000}$ and $W_{\rm out} \in \mathbb{R}^{1000 \times 50}$, and the output is read back out by the transpose $W_E^\top$.
Because $1000 > 100$ and $W_E$ has full row rank, the embedding is information-lossless: the network is equivalent to one acting directly on the input features through the effective weights $\widetilde W_{\rm in} = W_{\rm in} W_E^\top$ and $\widetilde W_{\rm out} = W_E W_{\rm out}$, which play exactly the roles of the encoder $W_{\rm in}$ and decoder $W_{\rm out}$ analyzed in \cref{sec:evidence,sec:open}.
It should therefore make no difference to the learned solution, and we confirm below that it does not.
\Cref{fig:embed-arch} shows this embedded architecture, with the fixed embedding $W_E$ and unembedding $W_E^\top$ restored relative to the axis-aligned model of \cref{fig:architecture}.

We retain the embedded variant because the compressed-computation toy model's original purpose was as a testbed for parameter-decomposition methods such as APD~\cite{braun2025apd} and SPD~\cite{bushnaq2025spd}, for which the non-axis-aligned features of the embedded model are the more challenging and realistic setting.
For the narrower claim that the trained network computes in superposition, the axis-aligned model in the main text is cleaner: with axis-aligned input features there is no random embedding that could, even in principle, be credited with the network's performance over the non-superposition baselines.

\subsection{The embedded model reproduces the axis-aligned solution}
\label{app:embed-match}

\begin{figure}[ht]
  \centering
  \includegraphics[width=\linewidth]{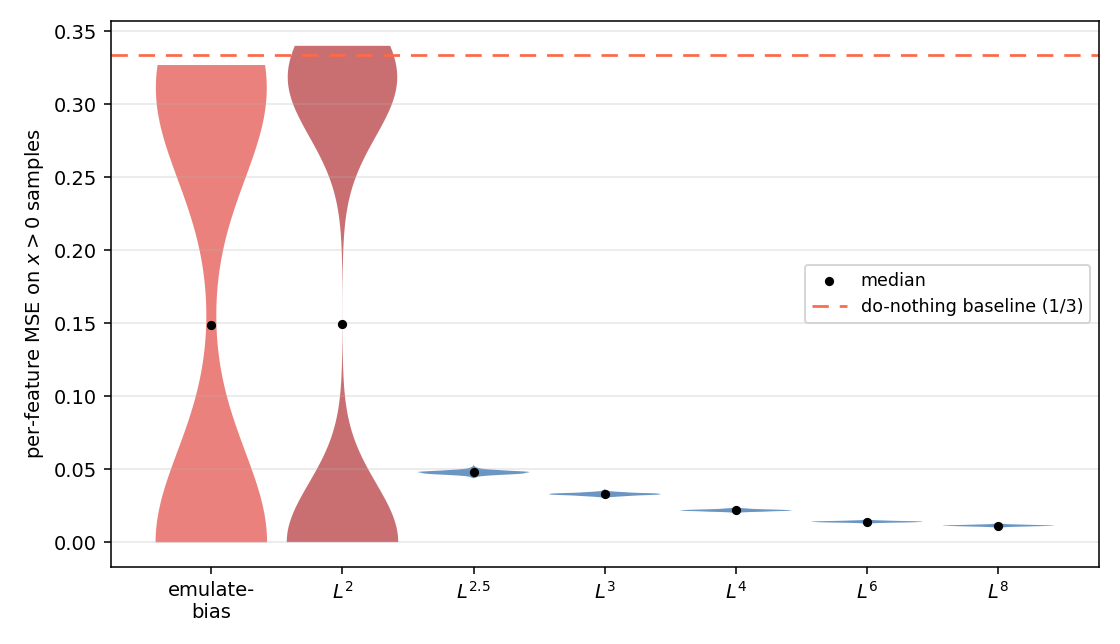}
  \caption{\textbf{Embedded model, per-feature MSE} (cf.\ \cref{fig:per-feature-loss}). As in the axis-aligned model, the emulate-bias baseline and $L^2$ are bimodal (half the features handled, half left near the do-nothing baseline) while every exponent above $2$ spreads the error evenly across all $100$ features.}
  \label{fig:embed-perfeature}
\end{figure}

\begin{figure}[ht]
  \centering
  \includegraphics[width=\linewidth]{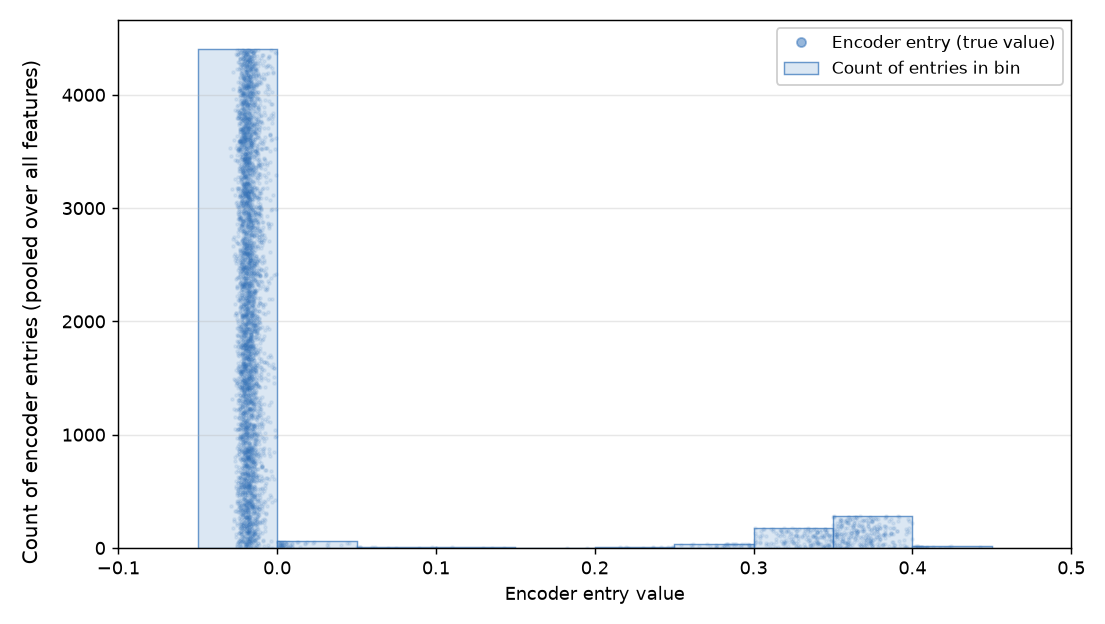}
  \caption{\textbf{Embedded model, effective-encoder entries} (cf.\ \cref{fig:value-hist}). Pooled histogram of the entries of the effective encoder $\widetilde W_{\rm in} = W_{\rm in} W_E^\top$; the same bimodal split into a small off-code cluster and a larger on-code cluster appears.}
  \label{fig:embed-value-hist}
\end{figure}

\begin{figure}[ht]
  \centering
  \includegraphics[width=\linewidth]{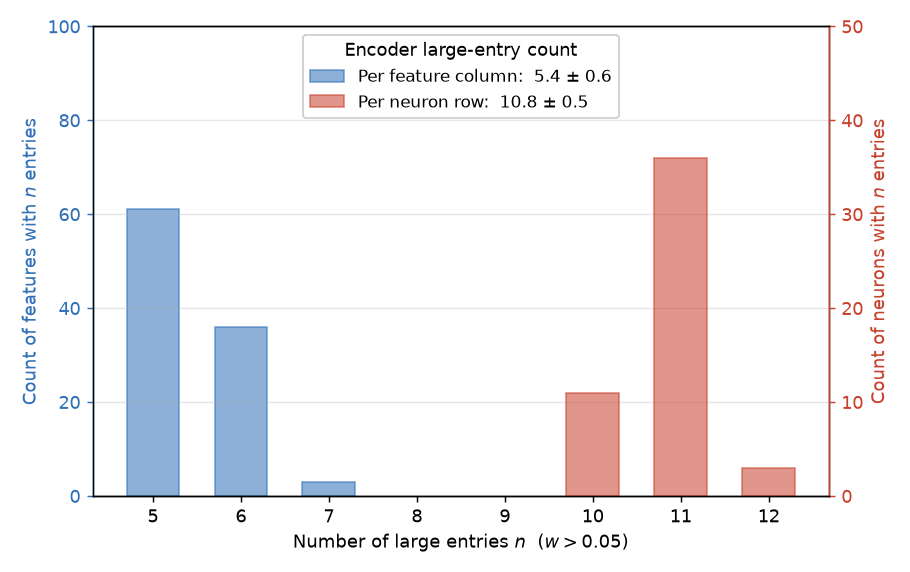}
  \caption{\textbf{Embedded model, codeword regularity} (cf.\ \cref{fig:code-hist}). Large-entry counts per feature column and per neuron row for the effective encoder; the code is close to biregular, as in the axis-aligned model.}
  \label{fig:embed-code-hist}
\end{figure}

We verify directly that adding the embedding changes nothing of substance.
We train the embedded model under the same loss, optimizer, and schedule, and recompute every quantity from \cref{sec:evidence,sec:open}.
\Cref{tab:embed} reports the comparison.
Across all of them --- the per-feature loss spread, the codeword-length distribution, the swap test, the decoder's agreement with a pseudoinverse, and the three-scalar description --- the two models are quantitatively the same; even the three fitted ansatz scalars $(a, b, c)$ agree to within $1\%$.
We conclude that the random embedding is incidental to the computation-in-superposition solution.
\Cref{fig:embed-perfeature,fig:embed-value-hist,fig:embed-code-hist} reproduce the main-text figures (\cref{fig:per-feature-loss,fig:value-hist,fig:code-hist}) for the embedded model.

\begin{table}[ht]
  \centering
  \small
  \setlength{\tabcolsep}{4pt}
  \caption{\textbf{The solution is unchanged by adding the random embedding.} Headline quantities from \cref{sec:evidence,sec:open} for the axis-aligned model (main text) and an otherwise identical model trained with \citet{braun2025apd}'s random embedding. CV is the coefficient of variation of the per-feature MSE across the $100$ features. The ansatz ratio is the $L^4$ loss of the three-scalar description relative to the trained model.}
  \label{tab:embed}
  \begin{tabular}{lcc}
    \toprule
    Quantity & Axis-aligned & Embedded \\
    \midrule
    Per-feature MSE CV ($L^4$) & $0.031$ & $0.031$ \\
    Per-feature MSE CV ($L^2$) & $0.999$ & $0.999$ \\
    Codeword length $K$ & $5\text{-}7$ & $5\text{-}7$ \\
    Swap-test pass rate & $100\%$ & $99.9\%$ \\
    Decoder--pinv cosine & $0.996$ & $0.996$ \\
    On/off-code value & $+0.34/{-}0.02$ & $+0.34/{-}0.02$ \\
    3-scalar ansatz ratio & $1.13\times$ & $1.13\times$ \\
    \bottomrule
  \end{tabular}
\end{table}

\subsection{Transpose versus pseudoinverse unembedding}
\label{app:pinv}

The embedded model reads its output back out with the transpose $W_E^\top$ of the embedding. Because $W_E$ has unit-norm but not orthogonal rows, this transpose is only an approximate inverse, leaving a small residual cross-talk between the near-orthogonal feature directions. We retrain the embedded model with its unembedding set instead to the exact pseudoinverse $W_E^{+}$ --- the left-inverse that removes this cross-talk --- holding the loss, optimizer, schedule, embedding, and seed fixed. It reaches the same $L^4$ loss (within $1\%$) and the same per-feature error spread.

\subsection{Other model sizes}
\label{app:sizes}

To check that the solution is not specific to our $F=100$, $N=50$, $d=1000$ choice, we vary the model size and measure the coefficient of variation of the per-feature MSE (\cref{sec:evidence}). Tightening the embedding to $d=50$ (keeping $F=100$, $N=50$) still gives indications of superposition, though weaker: the coefficient of variation is $0.20$, versus $0.03$ at $d=1000$. Scaling up an order of magnitude --- $F=1000$ features, a fixed random embedding into $d=500$, and $N=500$ neurons --- gives a low coefficient of variation again ($0.06$). In both regimes the error is spread across all features rather than concentrated as in the naive solution. We check only this loss statistic here, not the full binary-code mechanism, so we report these as indications of superposition rather than confirmations.

\section{Output attenuation}
\label{app:attenuation}

\begin{figure}[ht]
  \centering
  \includegraphics[width=\linewidth]{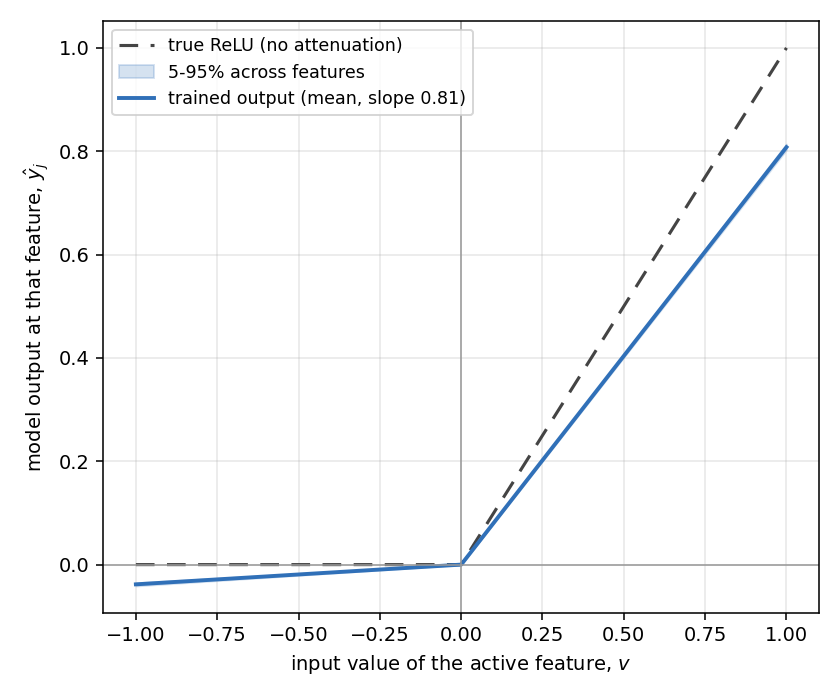}
  \caption{\textbf{Single-feature input--output response.} For each feature $j$ in turn, feature $j$ alone is driven with value $v$ and we read the output at index $j$; the blue curve is the mean over all $100$ features. The trained network traces a ReLU attenuated to mean slope $\approx 0.81$, below the true ReLU (dashed). The $5\text{--}95\%$ band across features is too narrow to be visible: every feature is attenuated almost identically, as expected for the even superposition solution.}
  \label{fig:io-response}
\end{figure}

Following \citet{bhagat2025compressed}, we examine the network's input--output response by driving a single feature $j$ with a value $v$ (all other features zero) and reading the output at index $j$, repeating this for each of the $100$ features (\cref{fig:io-response}). An un-attenuated network would trace $y = \mathrm{ReLU}(v)$; the trained network instead outputs a scaled-down ReLU, with mean slope $\approx 0.81$ on the active branch. The per-feature responses are nearly identical --- the $5\text{--}95\%$ band across features is too narrow to see --- consistent with the even, superposition-like solution in which every feature is handled the same way.

This attenuation is a response to interference. Because features are sparse (active with probability $p=0.02$) and represented by overlapping codewords, an active feature's signal leaks through shared neurons into the outputs of the inactive features, pulling those outputs above the zero they should produce. Under $L^4$ this cross-talk is costly. The network compensates by attenuating every active output: shrinking the on-feature response reduces the leaked interference on the far more numerous inactive features, at the cost of a small under-shoot on the active one. The net effect is the systematic under-shoot in \cref{fig:io-response}.

\section{Codeword overlap minimization}
\label{app:overlap}

When two features are active at once, their codewords interfere: neurons in both codewords receive contributions from both features, and the decoder must disentangle them.
The size of this interference is set by the pairwise overlap $|M_i \cap M_j|$, and under $L^4$ the largest overlaps dominate the loss.
Before fitting the ansatz to a synthetic code we therefore reduce its overlaps, and prefer codes whose overlaps are both small and uniform, which minimizes the worst-case interference.
We do this with a simple edge-swap procedure.

Let the pairwise overlap matrix be $O \in \mathbb{N}^{F \times F}$ with $O_{ij} = |M_i \cap M_j|$.
Our objective is the sum of off-diagonal squared overlaps, $\sum_{i \neq j} O_{ij}^2$.

At each iteration: sample two edges $(i, a), (j, b)$ uniformly at random from the edges in $M$; if $i \neq j$, $a \neq b$, $M_{ib} = 0$, and $M_{ja} = 0$, propose the swap removing $(i,a)$ and $(j,b)$ and adding $(i,b)$ and $(j,a)$; accept if the objective does not increase.

We run $8 \times 10^5$ iterations per code.
Edge swaps preserve codeword length (and, for biregular codes, neuron degree), so the structural properties of the family are kept fixed while the overlap distribution is optimized.

\end{document}